\documentclass[sigconf]{acmart}

\copyrightyear{2023}
\acmYear{2023}
\acmConference[CIKM '23]{Proceedings of the 32st ACM International Conference on Information and Knowledge Management}{October 21--25, 2023}{Birmingham, UK}
\acmDOI{}
\acmISBN{}

\AtBeginDocument{%
  \providecommand\BibTeX{{%
    \normalfont B\kern-0.5em{\scshape i\kern-0.25em b}\kern-0.8em\TeX}}}

\usepackage{float}
\usepackage{soul}
\usepackage{color}
\usepackage{bbm}
\usepackage{enumitem}

\definecolor{Blue}{rgb}{0,0,1}

\definecolor{Orange}{rgb}{1,0.5,0}

\definecolor{Green}{rgb}{0,1,0}

\begin{document}

\title[Tutorial]{Uplift Modeling: from Causal Inference to Personalization}

\author{Felipe Moraes}
\email{felipe.moraes@booking.com}
\affiliation{\institution{Booking.com}
\city{Amsterdam}
\country{Netherlands}}

\author{Hugo Manuel Proença}
\email{hugo.proenca@booking.com}
\affiliation{\institution{Booking.com}
\city{Amsterdam}
\country{Netherlands}}

\author{Anastasiia Kornilova}
\email{anastasiia.kornilova@booking.com}
\affiliation{\institution{Booking.com}
\city{Amsterdam}
\country{Netherlands}}

\author{Javier Albert}
\email{javier.albert@booking.com}
\affiliation{\institution{Booking.com}
\city{Tel Aviv}
\country{Israel}}

\author{Dmitri Goldenberg}
\email{dima.goldenberg@booking.com}
\affiliation{\institution{Booking.com}
\city{Tel Aviv}
\country{Israel}}

\begin{abstract}


Uplift modeling is a collection of machine learning techniques for estimating causal effects of a treatment at the individual or subgroup levels. Over the last years, causality and uplift modeling have become key trends in personalization at online e-commerce platforms, enabling the selection of the best treatment for each user in order to maximize the target business metric. Uplift modeling can be particularly useful for personalized promotional campaigns, where the potential benefit caused by a promotion needs to be weighed against the potential costs.
In this tutorial we will cover basic concepts of causality and introduce the audience to state-of-the-art techniques in uplift modeling.
We will discuss the advantages and the limitations of different approaches and dive into the unique setup of constrained uplift modeling.
Finally, we will present real-life applications and discuss challenges in implementing these models in production.

\end{abstract}



\maketitle

\section{Introduction}\label{sec:intro}
Uplift models~\cite{devriendt2018literature} are commonly used to estimate the expected causal effect of a treatment on the outcome of individuals, such as subscribing to a service, completing a purchase or responding to a medical treatment.
This can be achieved by estimating the \textit{Conditional Average Treatment Effect (CATE)}, defined as the expected increment in a user's outcome probability \textit{caused} by the treatment, given the individual's characteristics. 
It is particularly useful in the e-commerce setup, where we are interested in estimating the response to website changes for each of the users, in order to personalize their experience.

In the real world we can observe the outcome of a user only under the treatment she actually received and we will not know what would have happened, had she received a different treatment. As a result, we cannot directly calculate the treatment effect for any individual user. This problem, known as the fundamental problem of causal inference~\cite{holland1986statistics}, poses challenges in CATE estimation, since contrary to classical supervised machine learning, there is no labelled data available. 

Various CATE estimation techniques in the literature  try to overcome this problem in different ways, falling under two broad categories: \textit{meta-learners} and \textit{tailored methods}~\cite{zhang2020unified}. Meta-learning techniques, such as the two-models approach~\cite{hansotia2002direct}, the X-learner~\cite{kunzel2019metalearners} and outcome response transformations~\cite{athey2015machine,gubela2020response,proenca2023incremental} allow using classical machine learning techniques for estimating the CATE. Tailored methods, such as uplift trees~\cite{rzepakowski2012decision} and various neural network based approaches~\cite{johansson2016learning,louizos2017causal,yoon2018ganite}, modify well-known machine learning algorithms to be suitable for CATE estimation. 

Over the recent years uplift modeling has become popular in web and e-commerce applications, such as in Facebook, Amazon, Criteo, Uber and Booking.com \cite{makhijani2019lore, diemert2018large, zhao2019uplift, goldenberg2020free}
. In such applications, product improvements and promotions are typically tested via large-scale A/B testing, which allows estimating the overall treatment effect \cite{kaufman2017democratizing}. The data collected during the A/B tests can subsequently be used for uplift modeling, in order to distinguish between the \textit{voluntary buyers}, who would purchase even without receiving the treatment, and the \textit{persuadables}, who would only purchase as a response to the treatment~\cite{lai2006direct}. 

In the context of uplift modeling, an intriguing problem arises when the costs associated with different treatments vary among individuals \cite{haupt2022targeting,verbeke2023or}. In such scenarios, the objective often revolves around maximizing the overall incremental outcome while adhering to a total cost constraint. To address this, a constrained optimization problem can be solved by combining CATE estimations for both the outcome and the cost \cite{goldenberg2020free}.
Numerous recent studies have approached the issue of treatment allocation using various strategies. These include estimating the marginal uplift \cite{zhao2019unified}, incorporating net value optimization within meta-learners \cite{zhao2019uplift}, exploring the causal effect through bandits \cite{lin2017monetary}, modeling the optimization task as a min-flow problem \cite{makhijani2019lore}, and treating the problem as an online multiple-choice knapsack formulation \cite{albert2022commerce,yan2023end}.

Implementing uplift models in production environments gives rise to several operational challenges. For instance, models that are trained offline might be biased towards historical data and require dynamic calibration \cite{zhou2008budget} according to long-term changes and seasonality trends, which are particularly common in the travel industry \cite{mavridis2020beyond} and in promotional campaigns in general \cite{kim2006pay}. Another challenge is understanding and trust. It can be solved by providing global and local explainability \cite{kornilova2021mining}. This makes bridging between the underlying theory and practical implementations peculiarly relevant to applied data science research.

\section{Tutorial Outline}

The tutorial introduces key concepts on causality as well as recent advances in uplift modeling. The outline of the tutorial is as follows: First, we introduce basic concepts in causal inference under the Potential Outcomes framework. We continue with an overview of state-of-the-art uplift modeling techniques for evaluating and estimating conditional average treatment effects. Next, we discuss constrained uplift problems, a recent addition to the uplift modeling literature aimed at enabling cost-aware personalized treatment assignment. Lastly, we present real-life applications of uplift modeling and discuss challenges in implementing these models in production. The total duration of the tutorial is three hours.

\subsection{Detailed Schedule}

\begin{enumerate}
    \item Introduction to Causality (40 min)
    \begin{itemize}
        \item Potential Outcomes Framework
        \item Average Treatment Effects 
        \item Identifiability of Causal Effects
        \item Conditional Average Treatment Effects (CATE) 
    \end{itemize} 
    \item Uplift Modeling (60 min)
    \begin{itemize}
        \item Techniques for CATE Estimation 
        \begin{itemize}
            \item Meta-Learners
            \item Tailored Methods
        \end{itemize}
        \item Evaluating Uplift Models 
    \end{itemize}
    \item Uplift Modeling with Cost Optimization  (40 min)
    \begin{itemize}
        \item Types of Costs and Return-On-Investment
        \item Constrained Optimization
    \end{itemize}
    \item Applications and Implementation Challenges (40 min)
        \begin{itemize}
        \item Application Examples
        \item Model Robustness
        \item Exploration
        \item Adaptiveness 
        \item Explainability
    \end{itemize}
    
\end{enumerate}
 
\section{Relevance to the community}

In the past, causal inference has been associated mostly with clinical trials and social science applications. However, over the recent years, web applications have become increasingly more interactive, raising the need to estimate causal effects of online interventions to advance from correlation-based models to causal models. Learning the effects of different interventions and adjusting the personalization strategy accordingly has become a key trend in the e-commerce industry.

\subsection{Intended Audience}
The tutorial is targeted to industry practitioners and empirical researchers who are interested in getting causal insights from observational or interventional data and/or in building personalized e-commerce applications. 
As prerequisites, basic knowledge of probability, statistics and machine learning is expected. No prior knowledge of causal inference is required.

\subsection{Related Tutorials}

The tutorial is built upon materials from Booking.com's internal causal inference and machine learning trainings. Parts of the tutorial cover novel materials on recent research papers, talks, other tutorials and practical implementations. Throughout the tutorial we aim to convey our experience from a wide usage of uplift modeling in personalization applications at Booking.com.
In our recent tutorials at WSDM 2021 and WebConf'21~\cite{booking2021personalization, teinemaa2021uplift}, we present key trends in personalization, including an extended chapter on causality and uplift modeling. The proposed tutorial is intended to deep dive into uplift modeling topic, allowing extensive theoretical review and wide coverage of practical applications.

In prior tutorials by Kiciman and Sharma  at WSDM 2019~\cite{kiciman2019causal} and CoDS COMAD 2020-2023~\cite{sharma2020causal, karmakar2023causal}, the authors cover the importance of causality and pitfalls in conventional machine learning techniques that rely on correlation analysis. They present the concept of counterfactual reasoning and dive into methods for causal inference on large-scale online data. These methods set the ground for uplift modeling applications, providing a controlled training dataset for machine learning solutions. 

Another tutorial by Cui et al.~\cite{cui2020causal} at KDD 2020, introduces the strong relationship between causality and machine learning and describe various techniques for treatment effect estimation with advanced learning methods. Our proposed tutorial will cover similar techniques and expand on their applications in uplift modeling and treatment allocation optimization.

Besides these past tutorials, we rely on recent surveys by Devriendt et. al~\cite{devriendt2018literature}, Olaya et. al~\cite{olaya2020survey} and  Zhang et. al~\cite{zhang2020unified} that review state-of-the-art uplift modeling techniques and evaluation methods and compare their relative performance.

\subsection{Tutorial Format}

The suggested tutorial is a lecture-style tutorial, covering theoretical and practical aspects of uplift modeling.
Tutorial participants will be provided with supplemental materials, including tutorial slides, references to relevant literature and open-source libraries for uplift modeling.
The tutorial is build upon Booking.com internal training which is conducted online and in-person. The international team of presenters will give a special attention to interactive discussions during the session, taking into account the challenges of introducing the topic to a diverse audience.

\section{Presenters' Biography}

\textbf{Felipe Moraes} is a Machine Learning Scientist II at Booking.com in Amsterdam. He obtained his PhD on the topic of collaborative information systems from the Delft University of Technology. His current work is focused on uplift modeling and personalized marketing campaigns on the search results page. He has published his work in top venues, such as, SIGIR, CIKM, CHIIR, and ECIR.

\textbf{Hugo Manuel Proença} is a Senior Machine Learning Scientist I at Booking.com in Amsterdam. He obtained his PhD on the topic of interpretable machine learning models with statistical guarantees from the Leiden University. He has published his work in top conferences and journals like KDD, Information Sciences, and ECML-PKDD. His current interests focus on causal inference for personalization at scale. 

\textbf{Anastasiia Kornilova} is a Senior Machine Learning Scientist II at Booking.com in Amsterdam. She obtained her Masters in Applied Mathematics (with honors) from Ternopil National University. Her work was presented and published in top conferences like NeurIPS and WSDM. Her current work is focused on large-scale marketing campaigns optimisation and ML explainability.

\textbf{Javier Albert} is a Machine Learning Manager at Booking.com in Tel Aviv, where he uses uplift modeling to enable the financial viability of large-scale promotional campaigns. His work was recently published in top venues, such as, RecSys, WebConf and CIKM. Albert has a M.Sc. in Electrical Engineering from Tel Aviv University and two B.Sc. degrees from Technion - Israel Institute of Technology.

\textbf{Dmitri (Dima) Goldenberg} is a Senior Machine Learning Manager at Booking.com, Tel Aviv, where he leads machine learning efforts in recommendations, pricing and promotions personalization, utilizing online learning and uplift modeling techniques. Goldenberg obtained his Masters in Industrial Engineering and Management (with honors) from Tel Aviv University. He led the WSDM '21 and WebConf '21 tutorials on personalization and causal uplift modeling, and co-organized the WSDM '21 WebTour, KDD'22 WAMLM and Recsys'22 RecTour workshops. 
His research and applied work was presented and published in top journals and conferences including WebConf, CIKM, WSDM, SIGIR, KDD and RecSys.

\bibliographystyle{ACM-Reference-Format}
\bibliography{sample-base}

\end{document}